\documentclass[10pt,twocolumn,letterpaper]{article}

\usepackage[]{cvpr} 

\usepackage{amssymb}
\usepackage{pifont}
\usepackage{booktabs}
\usepackage{adjustbox}
\usepackage{float}
\usepackage{booktabs}
\usepackage{multirow}
\usepackage{siunitx}
\usepackage{xcolor}
\usepackage{caption}
\usepackage{mdframed}
\usepackage{enumitem}
\usepackage[accsupp]{axessibility}  

\usepackage{amsmath}   
\usepackage{booktabs}  
\usepackage{siunitx}   

\newcommand{\cmark}{\ding{51}}
\newcommand{\xmark}{\ding{55}}

\definecolor{cvprblue}{rgb}{0.21,0.49,0.74}
\usepackage[pagebackref,breaklinks,colorlinks,allcolors=cvprblue]{hyperref}

\def\paperID{35983} 
\def\confName{CVPR}
\def\confYear{2026}

\title{PinPoint: Evaluation of Composed Image Retrieval with Explicit Negatives, Multi-Image Queries, and Paraphrase Testing}

\author{Rohan Mahadev \quad Joyce Yuan \quad Patrick Poirson \quad David Xue \quad Hao-Yu Wu \quad Dmitry Kislyuk\\
Pinterest\\
{\tt\small \{rmahadev, jyuan, ppoirson, dxue, rexwu, dkislyuk\}@pinterest.com}
}

\begin{document}
\maketitle
\begin{abstract}
Composed Image Retrieval (CIR) has made significant progress, yet current benchmarks are limited to single ground-truth answers and lack the annotations needed to evaluate false positive avoidance, robustness and multi-image reasoning. We present \textbf{PinPoint}, a comprehensive real world benchmark with 7,635 queries and 329K relevance judgments all human verified across 23 query categories. PinPoint advances the field by providing: (1) multiple correct answers (averaging 9.1 per query) (2) explicit hard negatives, (3) six instruction paraphrases per query for robustness testing, (4) multi-image composition support (13.4\% of queries), and (5) demographic metadata for fairness evaluation. 

Based on our analysis of 20+ methods across 4 different major paradigms, we uncover three significant drawbacks: The best methods while achieving mAP@10 of 28.5\%, still retrieves irrelevant results (hard negatives) 9\% of the time. The best models also exhibit 25.1\% performance variation across paraphrases, indicating significant potential for enhancing current CIR techniques. Multi-image queries performs 40 to 70\% worse across different methods. To overcome these new issues uncovered by our evaluation framework, we propose a training-free reranking method based on an off-the-shelf MLLM that can be applied to any existing system to bridge the gap. We release the complete dataset, including all images, queries, annotations, retrieval index, and benchmarking code.
\end{abstract}

\section{Introduction}
\label{sec:intro}

\begin{figure}[h!]
  \centering
    \includegraphics[width=\linewidth]{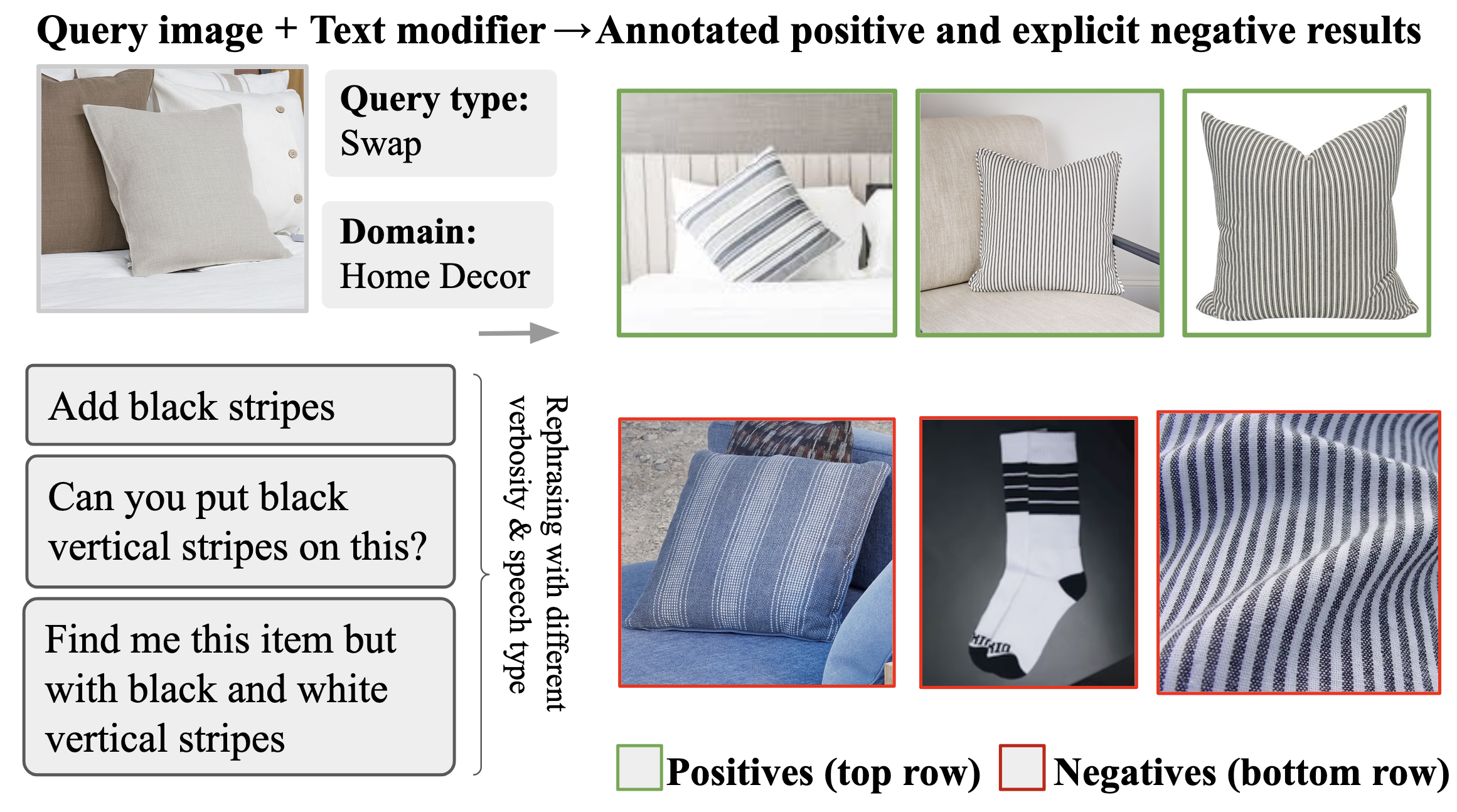}
    \caption{Example single image query from PinPoint demonstrating multiple instruction paraphrases, multiple ground truths (green), and explicit hard negatives (red)}
    \label{fig:single-ex}
\end{figure}


Visual search \cite{swain1991color, lowe2004distinctive} has evolved beyond image-to-image retrieval. With advances in multi-modal models \cite{radford2021learning, jia2021scaling}, composed image retrieval (CIR) \cite{vo2019composing,baldrati2023composed} enables users to combine reference images with natural language modifications. For example, someone redesigning a living room might reference one interior's color palette, another's furniture layout, and add ``make it more minimalist.'' This paradigm offers flexible search for applications from e-commerce \cite{ak2018learning, zhu2024bringing} to design inspiration \cite{zhai2017visual}.


Zero-shot CIR (ZS-CIR) \cite{baldrati2023zero, Saito_2023_CVPR} has become the standard protocol because it reflects real-world deployment constraints: models must generalize without task-specific fine-tuning. Two paradigms have emerged: (1) methods that compose modalities from foundation models (e.g., CLIP on LAION-400M \cite{schuhmann2021laion}) using learned combiners or projections, and (2) approaches that train directly on internet-scale triplets of (reference image, modification text, target image) assembled from web data \cite{zhang2024magiclens, zhou2025megapairs}. Both are evaluated zero-shot on held-out benchmarks to test true generalization.

\begin{figure}[htbp]
  \centering
  \includegraphics[width=0.95\columnwidth]{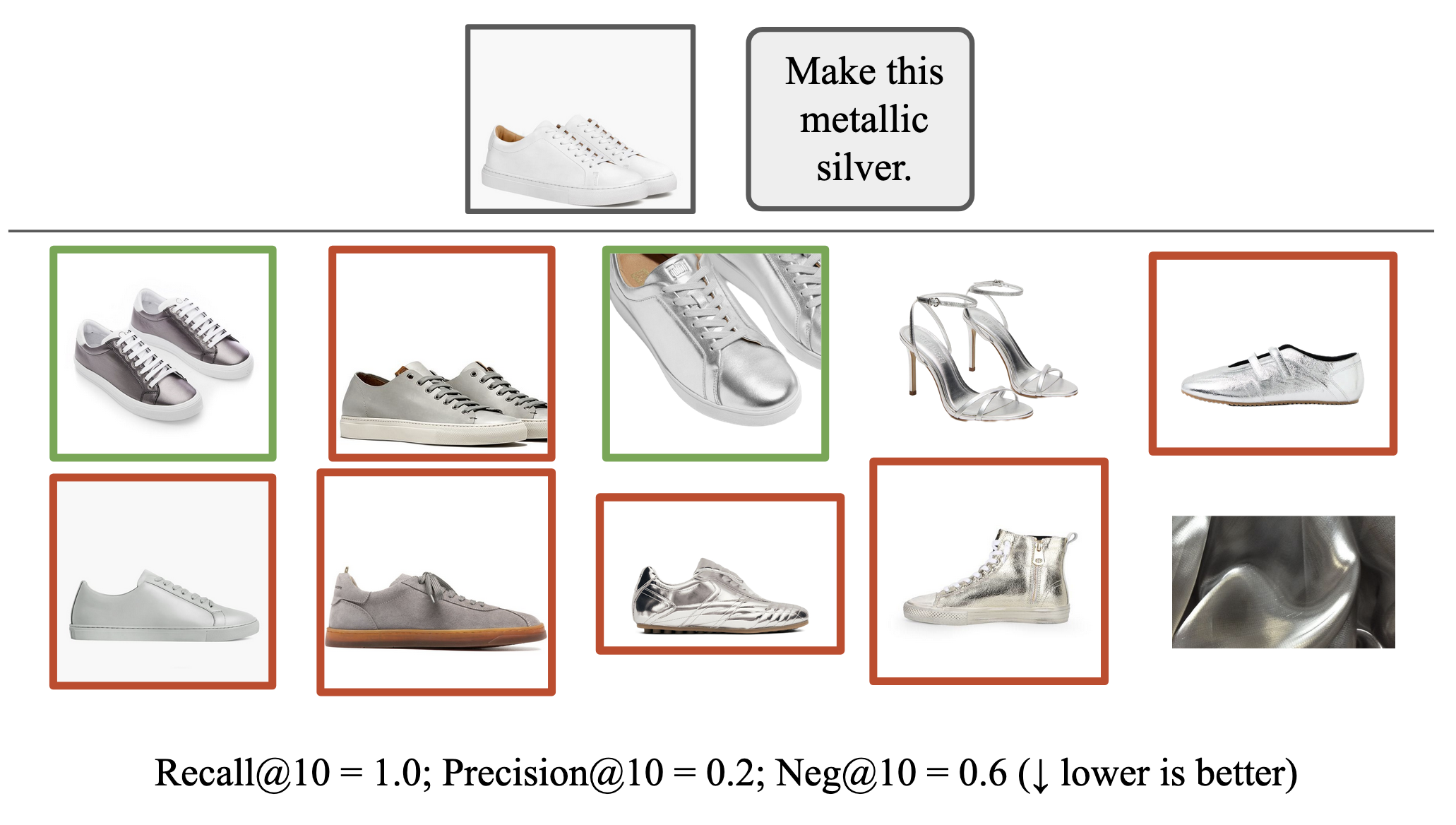}
  \caption{Metric pitfall:  Recall@10 = 1.0 yet 8 / 10 results violate the colour/material constraint (Precision@10 = 0.20, Neg@10 = 0.60).}
  \label{fig:cirr}
\end{figure}

Existing ZS-CIR benchmarks (CIRR and FashionIQ \cite{liu2021image, wu2021fashion}), have driven progress but exhibit fundamental limitations that misalign with real-world retrieval scenarios. First, they evaluate primarily through recall-based evaluation (does top-K contain any relevant image?), which ignores false positives. A system returning 2 relevant images and 8 distractors in top-10 scores identically to one returning 10 relevant results, illustrated in (\cref{fig:cirr}). Second, these benchmarks assume a single ground-truth answer per query. Yet as demonstrated by \citet{chun2025multiplicity}, multiplicity is inherent to multimodal matching - a single compositional query (``show me this dress in red'') may have dozens of valid matches with varying degrees of relevance. Ignoring this multiplicity means evaluation fails to capture ranking quality beyond the first hit.

The real-world search presents additional challenges absent from current benchmarks. Users may compose queries from multiple reference images (``outfit with [dress] and [shoes]''), testing whether models can reason compositionally across visual inputs. The same semantic intent can be expressed through varied phrasings (``show me this in red'' vs. ``change the color to red''), requiring robustness to linguistic variation. Search systems must also perform equitably across demographic groups and visual domains. While CIRCO \cite{baldrati2022effective} takes steps toward addressing some of these gaps, particularly explicit negatives and multi-domain coverage, its scale ($\approx$1000 queries) limits comprehensive evaluation of these scenarios.




We present PinPoint, a large-scale benchmark addressing these limitations. Built from real-world images spanning 23 diverse domains, PinPoint provides: (1) multiple annotated positives per query for measuring retrieval quality, (2) explicit hard negatives, visually similar distractors enabling direct false-positive measurement, (3) multi-image queries testing compositional reasoning, and (4) paraphrase variants and demographic metadata (based on Monk Skin Tone scale \cite{schumann2023consensus_monkmste}) for robustness and fairness analysis.

\begin{figure}[t]
  \centering
    \includegraphics[width=\linewidth]{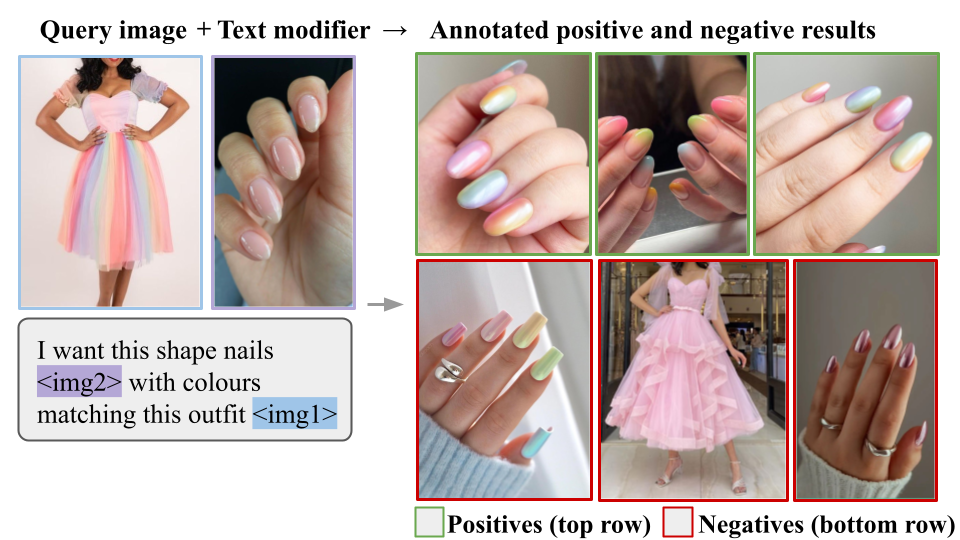}
    \caption{Multi-image composition query (13.4\% of PinPoint) requiring cross-image attribute extraction}
    \label{fig:multi-ex}
\end{figure}


Our evaluation of 20+ models spanning different CIR paradigms reveals substantial limitations invisible to existing benchmarks. All models exhibit substantially higher false-positive rates when evaluated against curated distractors, show dramatic sensitivity to instruction phrasing suggesting current techniques may overfit to benchmark patterns, and struggle significantly with multi-image queries. Interestingly, a text-only GPT \cite{openai2025gpt5} retrieval baseline outperforms many specialized CIR methods. We also propose a training-free reranking method using off-the-shelf MLLMs that consistently improves across all CIR methods. 

Our contributions can be summarized as follows:
\begin{itemize}
    \item \textbf{PinPoint benchmark}: A large-scale ZS-CIR evaluation dataset with 7,635 queries, 329K relevance judgments, explicit hard negatives, multiple ground truths (9.1 per query), multi-image composition support (13.4\% of queries), paraphrase variants (6 per query), and demographic metadata across 23 domains.

    \item \textbf{Comprehensive evaluation}: Analysis of 20+ models revealing critical limitations invisible to existing benchmarks, including high false-positive rates, sensitivity to instruction phrasing (25.1\% variation), and degraded multi-image reasoning.

    \item \textbf{Training-free reranking method}: A model-agnostic approach using off-the-shelf MLLMs that consistently improves performance across all CIR methods.

    \item \textbf{New evaluation protocols}: Metrics and analysis frameworks that account for multiplicity, explicit negatives, linguistic robustness, and demographic biases.
\end{itemize}

\section{Related Work}
\subsection{CIR Datasets and Benchmarks}
Current ZS-CIR benchmarks CIRR, CIRCO, FashionIQ, GeneCIS and MERIT establish important foundations but share critical limitations that hide real world failure modes.

CIRR (Composed Image Retrieval on Real‑life Images) \cite{Liu_2021_ICCV} was pioneering, with 36,554 queries derived from NLVR2 \cite{suhr2019corpus}, but it lacks evaluation dimensions: (i) no explicit negatives: showing nine wrong chairs before the correct one still yields perfect Recall@10; and (ii) instruction leakage, strong text encoders can retrieve the target without using the query image.


CIRCO \cite{agnolucci2025isearle} mitigates the single‑positive issue by allowing multiple positives, but still lacks explicit negatives, limiting robustness claims. FashionIQ \cite{wu2021fashion} focuses on fashion; while valuable for domain‑specific use, its narrow scope limits generalization and, like CIRR, it lacks explicit negatives and often assumes a single correct answer.


Newer datasets such as MERIT \cite{chow2025merit} make progress on interleaved multi‑image queries but remain domain‑limited. Several smaller datasets probe specific facets: GeneCIS \cite{Vaze_2023_CVPR} uses synthetic scene graphs; CLEVR\cite{johnson2017clevr}‑style datasets  offer controlled but unrealistic settings; Fashion200k \cite{han2017automatic} and Shoes \cite{NIPS2018_7348} are domain‑specific. None provide both multi‑image queries and comprehensive relevance labels needed for nuanced evaluation.

PinPoint operates at category level with instance-level hard negatives. Positives indicate semantic matches (e.g., ``red leather handbag''); negatives are visually similar distractors (e.g., ``red leather wallet''). This differs from \cite{psomas2025instance} which targets exact instance mathcing. We view these are complementary benchmarks.

\subsection{Major Families of CIR Methods}

Early methods composed separate image (CNN \cite{lecun1998convolutional}) and text (LSTM) features via residual gating \cite{vo2019composing}. With multimodal pretraining (e.g., CLIP, ALIGN), zero‑shot CIR became practical. Subsequent approaches include directly combining CLIP features \cite{baldrati2022effective}, mapping images to concept word tokens for text‑only composition \cite{saito2023pic2word}, and self‑masking projection enabling text‑only training \cite{gu2024language}. More recently, instruction tuning on curated datasets (e.g., MagicLens \cite{zhang2024magiclens}) and scaling such datasets yield state‑of‑the‑art results on common benchmarks \cite{zhou2025megapairs}.

\label{sec:bg-eval-primer}
Zero-shot CIR typically uses dual-encoder models (e.g., CLIP-style) that embed images and text in a shared space, enabling retrieval by similarity without task-specific training \cite{chuang2025meta,fang2023data}. At test time, the reference image and instruction can be combined via training-free fusion: (i) early fusion as a weighted sum of normalized image/text embeddings, or (ii) SLERP, which interpolates on the unit hypersphere between the two embeddings \cite{jang2024spherical}. For multi-image (compositional) queries, a simple and widely used baseline is mean pooling of reference-image embeddings when methods lack native composition support. Beyond direct encoders, proxy-based retrieval synthesizes a target \emph{proxy} conditioned on the image+instruction and then retrieves with a standard encoder: proxies can be text (detailed target descriptions; as shown by  \citet{yang2024ldre}) or images (edited/synthesized targets as shown by \citet{li2025imagine}).

\subsection{Post Retrieval Ranking}
A two-stage pipeline consisting of efficient first-stage retrieval followed by reranking has been used by \citet{liu2023candidate} for CIR and widely across other information retrieval tasks. LLM-based pointwise reranking for text-only queries has also been examined in Zhang \etal\cite{zhang2025qwen3} and Qu \etal \cite{qu2024unified}. MM-Embed \cite{lin2024mm} extends this approach to multimodal queries. This provides a robust training-free method for reranking which can be extended to ZS-CIR.
\section{PinPoint Dataset}

To address key shortcomings in existing CIR benchmarks, we introduce PinPoint, a composed image retrieval evaluation dataset targeting four gaps: (i) explicit negatives for false positive auditing, (ii) multi–image queries for compositional reasoning, (iii) paraphrase variants for linguistic robustness, and (iv) demographic/domain tags for subgroup analysis.

\begin{figure*}[t]
  \centering
  \includegraphics[width=\textwidth]{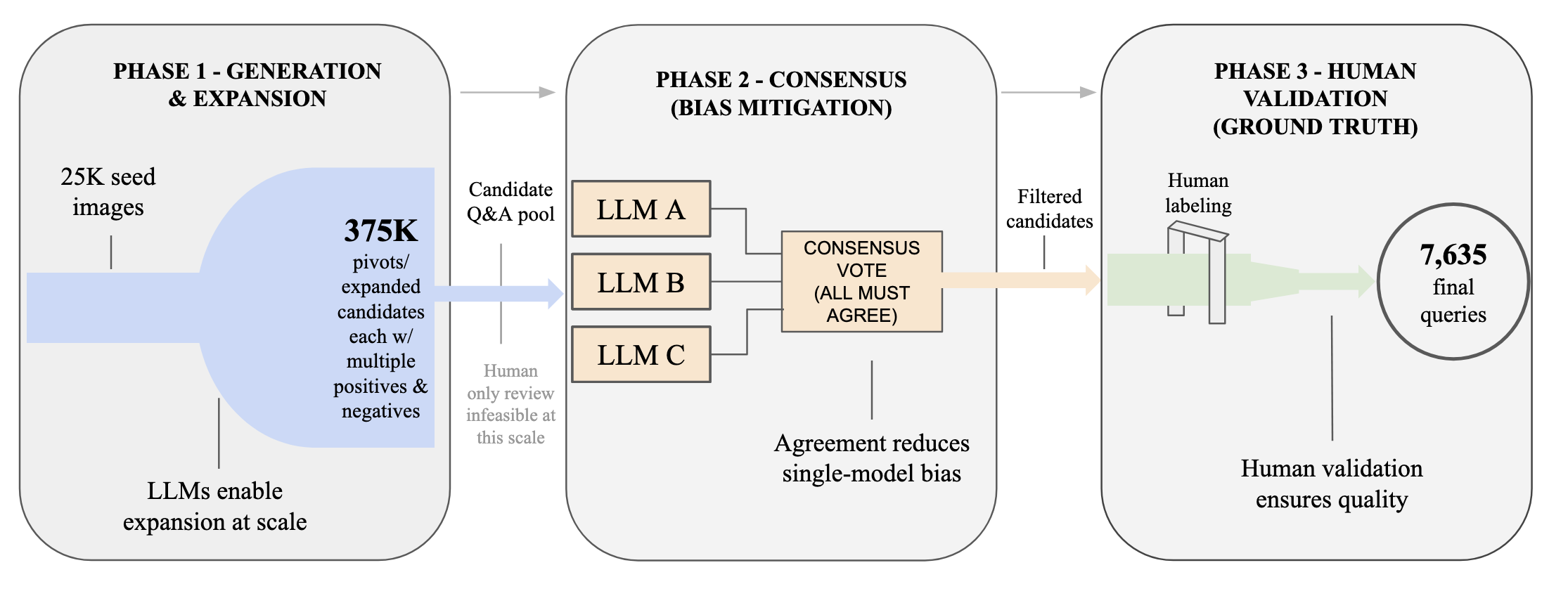}
  \caption{Dataset Construction Flow}
  \label{fig:construction}
\end{figure*}

\begin{table}[htbp]
\centering
\small
\setlength{\tabcolsep}{3pt}
\begin{tabular}{@{}lrcccc@{}}
\toprule
Benchmark & Queries & Multi- & Explicit & Multi- & Para- \\
 & /Index & Answer & Negatives & Image & phrases \\
\midrule
CIRR & 4K/2K & \xmark & \xmark & \xmark & \xmark \\
CIRCO & 800/123K & \cmark & \xmark & \xmark & \xmark \\
FashionIQ & 2K/5K & \xmark & \xmark & \xmark & \xmark \\
MERIT & 320K & \cmark & \xmark & \cmark & \xmark \\
\midrule
PinPoint & 8K/110K & \cmark & \cmark & \cmark & \cmark \\
\bottomrule
\end{tabular}
\caption{Comparison of CIR benchmarks}
\label{tab:benchmark-comparison}
\end{table}
\subsection{Dataset Construction}
Figure~\ref{fig:construction} overviews PinPoint’s construction pipeline. We sample 25,000 publicly available candidate query images scraped from the internet across 23 query categories (e.g., Fashion, Beauty, Home Decor, etc.). These images pass our automated quality filters (sufficient resolution, no blur/artifacts, visible primary objects) and safety screens, followed by de-duplication.

\subsubsection{Generating Diverse Modification Instructions}
For each query image we create 3--5 natural‑language modification instructions spanning five categories: Explore, Swap, Negation, Context Fit, and Complement. This captures common CIR intents beyond nearest‑neighbor matching.
For creating these instruction we use a two‑stage pipeline. First, we prompt three multimodal LLMs (GPT\mbox{‑}5, Claude 4 Sonnet, Gemini 2.5 Pro) \cite{openai2025gpt5,anthropic2025claude45,google2025gemini25}) each to propose five candidate instructions per image (15 total). Second, we de-duplicate and filter for specificity, visual grounding, topic alignment, and language quality; human raters then validate each generated instruction. For example, given a living‑room image, “make it more rustic with wooden furniture” is accepted, whereas “make it better” is rejected for vagueness.

\subsubsection{Paraphrase Generation for Robustness Evaluation} \label{sec:paraphrases}
Users express the same intent in many ways (e.g., “make it blue,” “change the color to blue”). We vary verbosity (concise vs. detailed) and speech type (imperative vs. questions). Each paraphrase shares the same positive and negative annotations, allowing us to measure consistency across linguistic variations. Each paraphrase was generated first by LLMs and then human verified for specificity, topic-alignment, diversity, language quality and visual grounding.

\subsubsection{Multi-Answer Annotation with Explicit Negatives} \label{sec:consensus}
A critical limitation of existing datasets is their assumption of a narrow range of correct answers (e.g., “change this shirt to blue” spans many blue shirts across poses and scenes). We aim to be exhaustive in our annotation of both positive and negative candidates. For each query, three multimodal LLMs propose short descriptors of valid targets and plausible false positives. For the example query ``change the shirt to blue'', correct descriptions might include ``navy blue button down shirt'', while incorrect ones might be ``red shirt'' (wrong color) or ``blue pants'' (wrong item). We scrape upto 50 candidates for each descriptor from the web, yielding 100 candidates per query. The same three LLMs then rate each candidate into 5 categories: \{\emph{Very Relevant, Somewhat Relevant, Irrelevant, Somewhat Irrelevant, False Positive}\}. We keep only unanimous \emph{Very Relevant} as positives and unanimous \emph{False Positive} as explicit negatives; human raters verify each accepted positive and negative. On average, this results in 9.1 positives and 32.8 explicit negatives per query (Table~\ref{tab:dataset-stats}). We then add all of these positives and negatives along with many irrelevant images to our retrieval corpus of 109,601 images.

\subsubsection{LLM Bias in Dataset Construction}
Our dataset construction implements 3 layers of safeguards against LLM bias:

\textbf{1. Human Verification}: Every final query, positives, and negative candidates are human-reviewed. Of the approx. 12,500 LLM proposed queries, human annotators rejected 37\% as ambiguous or incorrect. The 7,635 released queries survived this filter as the final decision on acceptance was made by humans.

\textbf{2. Multi-LLM Consensus}: We do NOT rely on a single model. GPT-5, Claude, and Gemini each independently generate candidates and during filtering only candidates where each LLM agrees on the acceptance proceed.

\textbf{3. Scale Necessity}: In order to create the diverse dataset with the scale of queries and number of positives and negatives, we need to employ the use of LLMs for practically doing so. LLMs enable this scale and the human review ensures the quality.

\begin{table}[htbp]
\centering
\caption{Dataset statistics for PinPoint}
\label{tab:dataset-stats}
\begin{tabular}{lr}
\toprule
Base queries & 7,635 \\
Images in corpus & 109,601 \\
Avg positives per query & 9.1 \\
Avg negatives per query & 32.8 \\
Multi-image queries & 13.4\% \\
Paraphrases per query & 6 \\
Question-format queries & 16.7\% \\
Domain categories & 23 \\
Demographic annotations & Monk Skin Tones \\
\bottomrule
\end{tabular}
\end{table}

\subsubsection{Distributions}
\begin{figure*}[t]
  \centering
    \includegraphics[width=\linewidth]{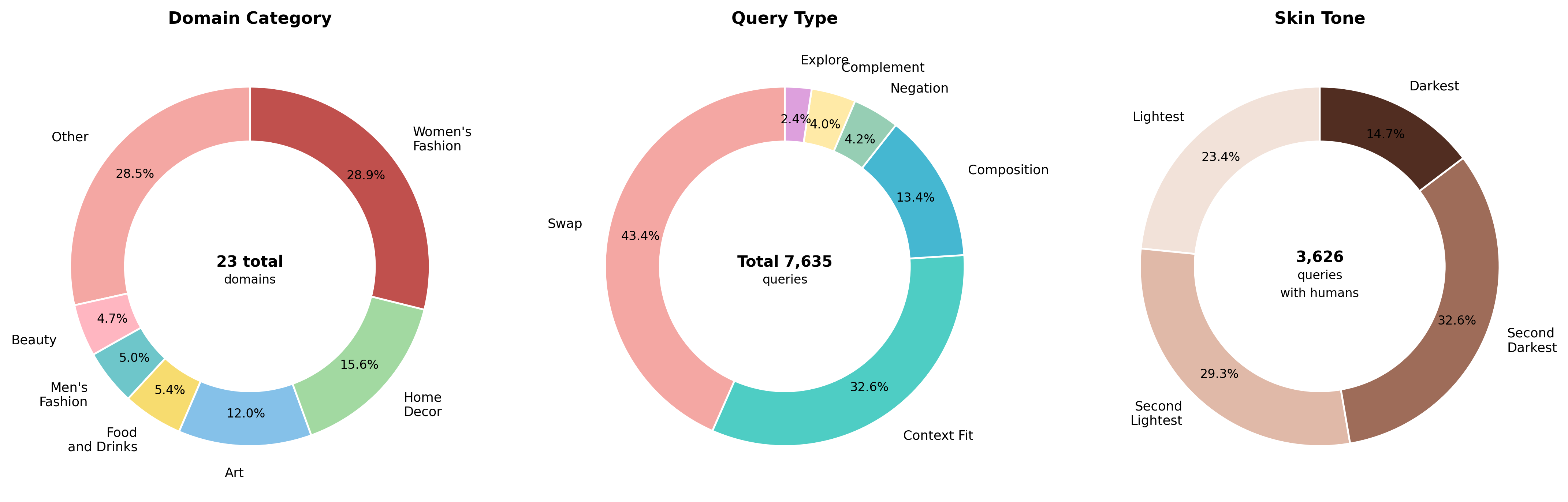}
\caption{PinPoint distributions. Left-to-right: (a) query domain categorization; (b) instruction type mix; (c) Skin Tone buckets for people-containing queries.}
  \label{fig:pp-distributions}
\end{figure*}

We check PinPoint’s composition with three distributions (Fig.~\ref{fig:pp-distributions}): (i) query domain, (ii) instruction type (Explore/Swap/Negation/Context Fit/Complement), and (iii) Monk Skin Tone \cite{Monk_2019} ranges for query images. These summarize the dataset properties used later for subgroup and robustness analyses.




\section{Evaluation}
\subsection{Metrics}

We look at standard retrieval metrics like mean average precision (mAP) as a benchmark metric. But, given PinPoint's added multi-answer annotations and explicit negatives, we evaluate several performance indicators in addition to standard metrics.
\subsubsection{$\Delta$mAP@10}
 With the addition of the explicit negatives in the dataset, we want to evaluate the impact these false positives add to the performance of models. Therefore, we compute
\begin{equation}
\Delta\text{mAP@10} = \text{mAP@10}_{\text{no\_hn}} - \text{mAP@10}_{\text{all}}
\label{eq:1}
\end{equation}
where $\text{mAP@10}_{\text{no\_hn}}$ is the mean average precision@10 computed without hard negatives in the retrieval corpus, and $\text{mAP@10}_{\text{all}}$ is computed with hard negatives included.
This metric indicates the tendency of each method to retrieve false positives. Robust models are expected to exhibit a $\Delta\text{mAP@10}$ score close to zero, indicating minimal sensitivity to the inclusion of false positives. We also report the Negative Recall@10, which is the frequency of false positives in the top-10 results.

\subsubsection{Linguistic Sensitivity}
With the six paraphrases for each query in the dataset, we can measure the sensitivity of each method to these paraphrases as:
\begin{equation}
\text{sensitivity\_range} = \max_{p \in \mathcal{P}_q}\left(\text{mAP@10}_p\right) - \min_{p \in \mathcal{P}_q}\left(\text{mAP@10}_p\right)
\end{equation}

where $P_q$ represents the 6 paraphrases for query $q$, averaged over all queries.

Models showcasing lower sensitivity\_range scores indicate a higher robustness to the phrasing of queries. 






\subsection{Methods}\label{sec:evaluated_models}
We evaluate 20+ models across 4 paradigms in a zero-shot setting (no PinPoint-specific training).

\subsubsection{CLIP Baselines}
These include general vision–language encoders without CIR-specific training: Meta CLIP 2 \cite{chuang2025meta} and Apple DFN-CLIP \cite{fang2023data} (base/huge). For multimodal queries we report four inference recipes: (i) image-only, (ii) text-only, (iii) early fusion of normalized image/text embeddings, and (iv) SLERP interpolation \cite{jang2024spherical}.
\subsubsection{CIR Specific Methods}
We evaluate composed-image retrieval methods, namely MMRet \cite{zhou2025megapairs} (CLIP and MLLM-based variants), MagicLens (base and large), LinCIR \cite{gu2024language}, and Pic2Word \cite{saito2023pic2word}. For multi-image compositional queries, we average the reference-image embeddings (mean pooling) for methods lacking native composition support; otherwise, we adopt the authors’ recommended inference procedure.


\paragraph{Proxy-based Methods}
\label{sec:proxies}
As an alternative to direct encoders, we evaluate proxy-based retrieval by generating a target \emph{proxy} from the reference image and instruction, embedding the proxy, and retrieving candidates.

\paragraph{Text proxies}
We prompt a frontier LLM (GPT-5) to produce detailed target-image descriptions from the reference image and instruction, then perform text-only retrieval with Meta CLIP 2 text embeddings. We generate short/medium/long variants and merge the results in two ways: pre-merge where we average the text-embeddings of variants encoded by Meta CLIP 2's text encoder before retrieval, and post-merge where we combine results post-retrieval for each variant based on similarity score.

\subsubsection{Training-Free Point-wise Reranking}
To improve top-ranked precision and suppress false positives without retraining the base retriever, we introduce a training-free, point-wise reranking scheme based on a Multimodal Large Language Model (MLLM).

Given a query image $I_q$, an instruction text $T$, and a set of candidate images $\{I_c^{(i)}\}_{i=1}^{N}$ retrieved in the first stage, we compute relevance scores using Qwen2.5-VL-7B \cite{bai2025qwen2}. For each candidate $I_c^{(i)}$, we prompt the MLLM with the query image, the instruction, and the candidate image:

\begin{quote}
\small
\emph{``Given the query image and the instruction: \texttt{\{T\}}, is the candidate image relevant? Answer with `yes' or `no' only.''}
\end{quote}

The MLLM produces logits for the tokens ``yes'' and ``no'':

\begin{equation}
\ell_{\text{yes}}^{(i)}, \ell_{\text{no}}^{(i)} = \text{MLLM}(I_q, T, I_c^{(i)}),
\end{equation}

where $\ell_{\text{yes}}^{(i)}$ and $\ell_{\text{no}}^{(i)}$ are the raw logit values for the ``yes'' and ``no'' tokens, respectively.

We convert these logits into the probability of relevance via a sigmoid over the logit difference:
\begin{equation}
P(\text{relevant} \mid I_c^{(i)}) =
\sigma\!\left(\ell_{\text{yes}}^{(i)} - \ell_{\text{no}}^{(i)}\right)
\end{equation}
We take this probability as the final reranking score $s_{\text{rerank}}^{(i)}$ for candidate $i$. The candidates are finally sorted in descending order of $s_{\text{rerank}}^{(i)}$, yielding a drop-in reranking mechanism that can boost performance on top of any first-stage retriever without additional training.

\subsection{Implementation Details}
\label{sec:impl}
We run batched embedding inference on NVIDIA A100 GPUs and use FAISS \cite{douze2025faiss} with FlatIndexIP indexing for nearest-neighbor search. CIR-specialized models follow their published inference defaults. For reranking we use Qwen2.5-VL-7B via vLLM \cite{kwon2023efficient}.

\textbf{Compute and latency:} Embedding-based methods process queries in 0.02–0.15s each (model-size dependent). Generation-based approaches run sequentially: image generation takes 2–5s per query; text generation 1–2s. These overheads contextualize throughput differences versus direct embedding approaches. With KV-cache prefill (no autoregressive generation), latency for the reranker is around \textbf{120ms} per candidate on a single GPU deployed using the VLLM framework.

\section{Results and Analysis}

\begin{table*}[t]
\centering
\caption{Performance comparison of different methods for composed image retrieval. Metrics include mAP@10, mAP@10 without explicit negatives, relative decline in mAP, Negative Recall at 10, and linguistic sensitivity range.}
\label{tab:fullresults}
\resizebox{\textwidth}{!}{%
\begin{tabular}{@{}l S[table-format=1.3] S[table-format=1.3] S[table-format=2.2, detect-weight=true] S[table-format=1.3] S[table-format=1.3]@{}}
\toprule
\textbf{Method} &
\multicolumn{1}{c}{\textbf{mAP@10}} &
\multicolumn{1}{c}{\textbf{mAP@10$_{\text{no-hn}}$}} &
\multicolumn{1}{c}{\textbf{$\Delta$mAP (\%)$\downarrow$}} &
\multicolumn{1}{c}{\textbf{NegRecall@10$\downarrow$}} &
\multicolumn{1}{c}{\textbf{Ling. Sens.$\downarrow$}} \\
\midrule
\multicolumn{6}{@{}l@{}}{\textit{\textbf{CLIP Family}}} \\
\quad Meta CLIP 2 -- Image Only & 0.017 & 0.030 & 77.09 & 0.164 & {\textbf{0.037}} \\
\quad Meta CLIP 2 -- Text Only & 0.031 & 0.033 & 4.95 & 0.018 & 0.087 \\
\quad Meta CLIP 2 -- SLERP & 0.043 & 0.065 & 48.78 & 0.084 & 0.115 \\
\quad Meta CLIP 2 -- Combined & 0.044 & 0.062 & 39.87 & 0.072 & 0.114 \\
\midrule
\multicolumn{6}{@{}l@{}}{\textit{\textbf{CIR Family}}} \\
\quad Pic2Word & 0.046 & 0.068 & 45.63 & 0.153 & 0.106 \\
\quad LinCIR & 0.110 & 0.135 & 23.47 & 0.141 & 0.152 \\
\quad MagicLens-CLIP-B & 0.111 & 0.128 & 15.78 & 0.159 & 0.152 \\
\quad MMRet-CLIP-B & 0.153 & 0.173 & 12.72 & 0.123 & 0.176 \\
\quad MagicLens-CLIP-L & 0.155 & 0.178 & 14.41 & 0.151 & 0.182 \\
\quad MMRet-CLIP-L & 0.178 & 0.197 & 10.89 & 0.120 & 0.188 \\
\quad MMRet-MLLM-S2 & 0.203 & 0.239 & 18.03 & 0.151 & 0.149 \\
\quad MMRet-MLLM-S1 & 0.224 & 0.238 & 6.38 & 0.091 & 0.162 \\
\midrule
\multicolumn{6}{@{}l@{}}{\textit{\textbf{Text Generation}}} \\
\quad GPT-5-Text -- Postmerge & 0.236 & 0.255 & 7.94 & 0.094 & 0.180 \\
\quad GPT-5-Text -- Premerge &\textbf{0.266} & {\textbf{0.285}} & 6.93 & 0.090 & 0.174 \\
\midrule
\midrule
\multicolumn{6}{@{}l@{}}{\textit{\textbf{with Reranking}}} \\
\quad Meta CLIP 2 -- Combined + Reranking & 0.087 & 0.089 & 2.71 & 0.039 & 0.149 \\
\quad MMRet-CLIP-L + Reranking & 0.236 & 0.244 & 3.45 & 0.074 & 0.208 \\
\quad MagicLens-CLIP-L + Reranking & 0.231 & 0.238 & 3.14 & 0.078 & 0.201 \\
\quad GPT-5-Text -- Premerge + Reranking & 0.272 & 0.281 & 3.46 & 0.062 & 0.193 \\
\quad MMRet-MLLM-S1 + Reranking & {\textbf{0.290}} & {\textbf{0.295}} & {\textbf{2.01}} & {\textbf{0.056}} & 0.191 \\
\bottomrule
\end{tabular}%
}
\end{table*}

\subsection{False Positive Retrieval Tendencies}
\cref{tab:fullresults} shows the $\Delta$mAP as defined in \cref{eq:1}. We observe worse performance of all models when explicit negatives are added to the corpus. This highlights a clear gap in existing evaluation benchmarks, and how state-of-the-art ZS-CIR models are prone to retrieving false positives, which doesn't show the same improvement trends as compared to regular mAP. We therefore need further interventions in these models to bridge the gap between mAP with and without explicit negatives.

\subsection{Precision-Safety Trade off}
\begin{figure}[htbp]
  \centering
  \includegraphics[width=\columnwidth]{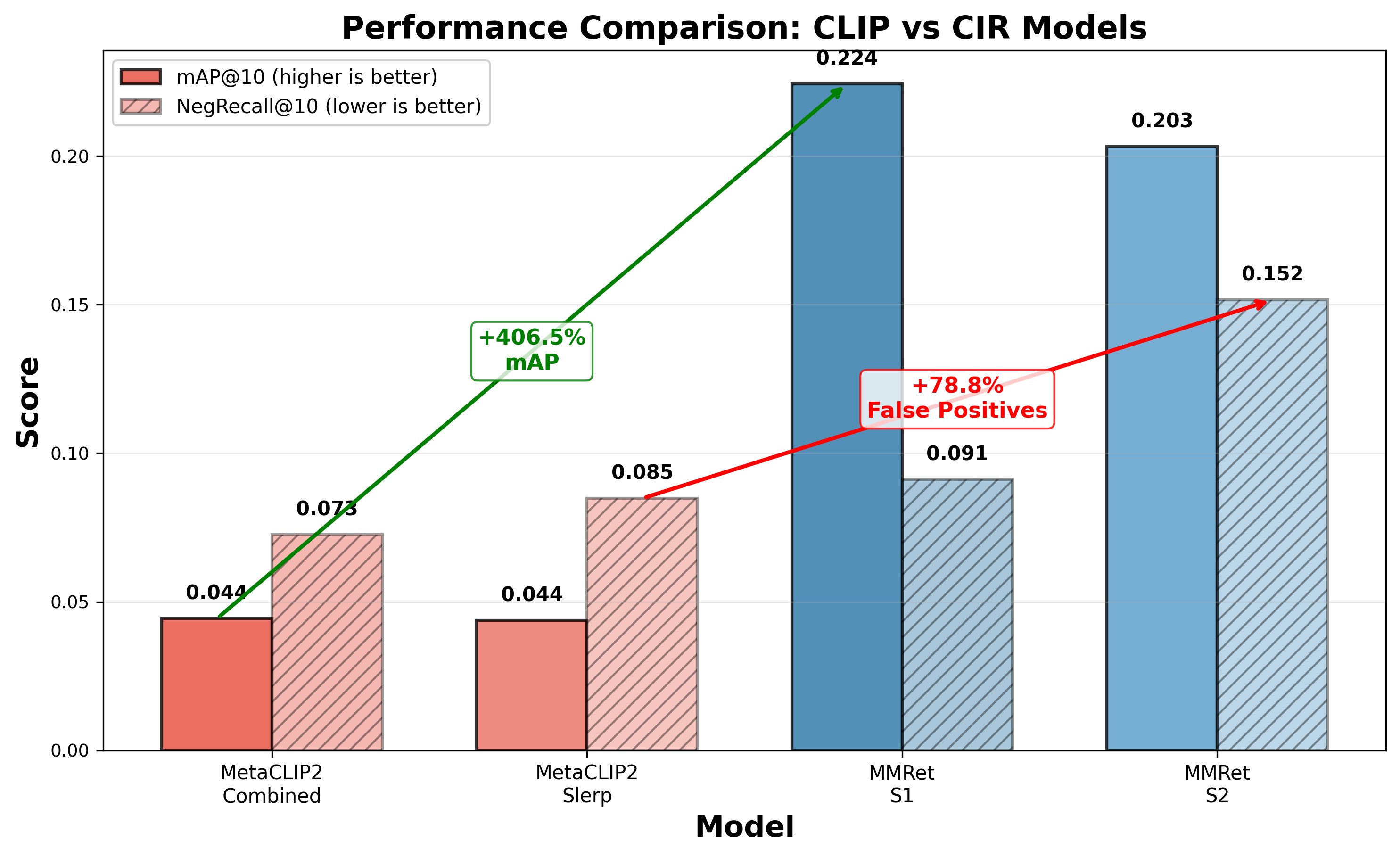}
  \caption{Performance comparison showing CIR models achieve better mAP but worse negative recall compared to CLIP baselines}
  \label{fig:precision-safety}
\end{figure}

\cref{fig:precision-safety} directly compares CLIP baselines against CIR-specialized models, measuring both mAP and negative suppression using Negative Recall @10. We see that MMRet-S1 achieves 3.4x better mAP than Meta CLIP 2 but retrieves 25\% more false positives. The metrics trend towards the bottom right, with imperfect trajectories - showing that we improve mAP but perform worse at Negative Recall with CIR models. This shows a need for better balancing this tradeoff as we improve on other benchmarks.

\subsection{Improvements with Reranker}
From \cref{tab:fullresults}, we observe that without MLLM based reranking, GPT-5 based text generation performs the best which is expected given its known abilities of complex reasoning and its relative compute size \cite{yue2024mmmu}, followed by state-of-the art CIR methods like MMRet. However,  when we rerank the results of MMRet-S1 model with the MLLM, we see that the performance overtakes that of GPT-5 on both mAP@10 but also negative avoidance as measured by Negative Recall@10. This improvement in overall metrics can be seen across the board for all models, which shows the versatility of the training-free MLLM reranking approach and how we could reach state-of-the-art performance by using a combination of methods.

\begin{figure}[htbp]
  \centering
  \includegraphics[width=\columnwidth]{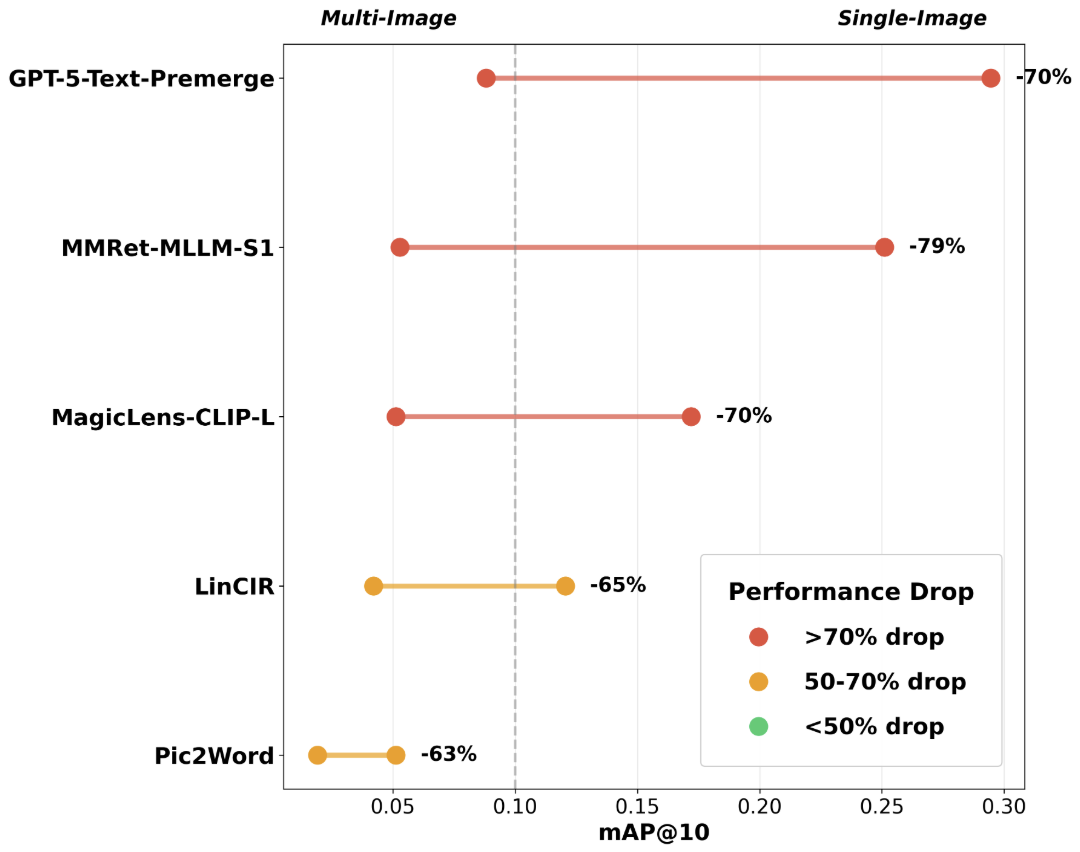}
  \caption{All models suffer 48-72\% performance drop on multi-image queries}
  \label{fig:singmul}
\end{figure}

\begin{figure}[htbp]
  \centering
  \includegraphics[width=\columnwidth]{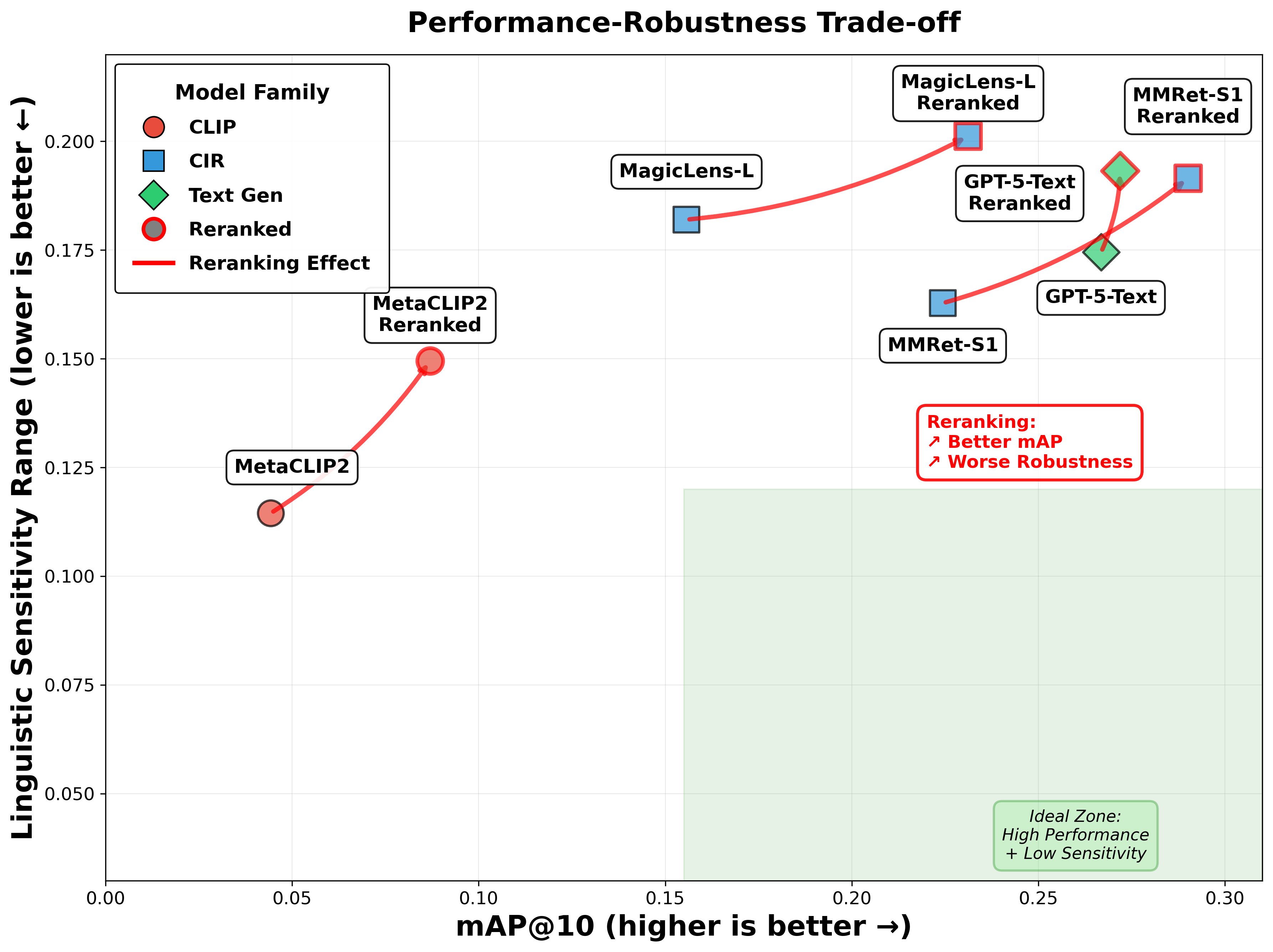}
  \caption{Linguistic Sensitivity worsens with improved mAP performance}
  \label{fig:ling_sens}
\end{figure}

\subsection{Multi-Image Queries}
\cref{fig:singmul} reveals ZS-CIR’s critical unsolved problem: all models suffer from poor performance on multi-image queries, dropping 48-72\% in performance between single-image and multi-image queries. The best absolute performance by MMRet-S1 only achieves 0.067 mAP@10 on multi-image queries with reranking, highlighting a persistent performance gap on multi-image queries. Even frontier LLMs don’t perform exceptionally well. \cref{tab:multi-image-perf} shows the difference in performance ratios. We recognize the shortcomings of the experimentation as we had

\begin{table}[htbp]
\centering
\small
\setlength{\tabcolsep}{4pt}
\begin{tabular}{lrrr}
\toprule
Method & Single & Multi & Ratio \\
\midrule
MMRet-MLLM-S1 & 0.324 & 0.067 & 4.83$\times$ \\
MMRet-MLLM-S2 & 0.227 & 0.051 & 4.45$\times$ \\
MMRet-CLIP-L & 0.262 & 0.063 & 4.15$\times$ \\
MagicLens-L & 0.257 & 0.062 & 4.14$\times$ \\
LinCIR & 0.121 & 0.042 & 2.88$\times$ \\
\midrule
MMRet-CLIP-L + Reranking & 0.262 & 0.063 & 4.15$\times$ \\
MMRet-MLLM-S1 + Reranking & 0.324 & 0.067 & 4.83$\times$ \\
MagicLens-L + Reranking & 0.257 & 0.062 & 4.14$\times$ \\
GPT-5-Text-Premerge + Reranking & 0.302 & 0.074 & 4.08$\times$ \\
\bottomrule
\end{tabular}
\caption{Single-image vs multi-image query mAP@10 performance}
\label{tab:multi-image-perf}
\end{table}

\subsection{Linguistic Sensitivity}
\cref{fig:ling_sens} reveals a critical paradox: High-performing models consistently show 3-5× higher linguistic sensitivity than baseline CLIP models, suggesting they overfit to specific phrasings rather than learning robust representations. Reranking, while boosting mAP@10 by 30-50\%, degrades rephrasing robustness by 10-30\%, with MagicLens-Large worsening from 0.182 to 0.201 (+10\%) and Meta CLIP 2 from 0.115 to 0.150 (+30\%) after reranking.

\section{Discussion}

\subsection{CIR Specialization is Essential}

Task-specific training, even on non-domain-specific datasets, provides dramatic gains. MMRet-MLLM-S1 achieves $+238\%$ better mAP compared to CLIP baselines (Meta CLIP 2), validating the need for composed retrieval architectures rather than generic vision-language models. However, this comes at a tradeoff of $+25\%$ increase in Negative Recall@10, suggesting current training paradigms prioritize positive matching while neglecting negative suppression.

\subsection{Rerankers as an intervention to false positive issues}

We show that a training-free MLLM based reranker can alleviate issues caused by the addition of the distractor negatives. From \cref{tab:fullresults}, we see consistent improvement in the mAP@10, regardless of the presence of explicit negatives in the dataset, suggesting that the reranker adds robustness to explicit negatives. More specialized training on ranking based datasets could potentially improve this performance even further.

\subsection{Shortcomings of Reranking}

\cref{tab:fullresults} also highlights the worsening of the linguistic sensitivity with the reranker across all models, which is a sign that the training-free MLLM based rerankers are not a catch-all fix to issues for CIR. Also \cref{tab:multi-image-perf} shows how multi-image performance does not improve with reranking. The reranker demonstrates that PinPoint's challenges are addressable, not solved. It shows that with minimal but targeted intervention, all 20+ tested models improve. Therefore, this suggests we need more foundational approaches to fix these issues, including training models with a linguistically diverse dataset and better architectures that compose image features in a more sophisticated manner.


\section{Limitations}

\textbf{Dataset Scope} Our 23 lifestyle-based domain categories provide broad coverage, but the dataset lacks specialized domains like industrial design, medical imaging or satellite imagery. Geographic and cultural biases exist as studied in \cite{shankar2017no}, with sampling skewed towards Western concepts and English-only queries, limiting global generalization. Aesthetic/style queries have inherent subjectivity, and $\sim$10 positive answers per query may not be fully exhaustive. Multi-image queries are fixed to two images, but real-world applications may compose 5+ images.\newline
\textbf{Generalization} We only present zero-shot evaluation results. Supervised fine-tuning on a dataset similar to PinPoint could arguably solve many of these problems, but due to costs associated with collecting such a dataset at training scale, we limit our current study to the evaluation of existing models. Future studies on how training datasets with multiple positives, explicit negatives, and interleaved multi-image queries would improve performance remain important directions.
\section{Conclusion}


In this work we introduce PinPoint, the first CIR benchmark with explicit negative annotations, multi-image queries, and paraphrase sensitivity testing. \textbf{Second} through comprehensive evaluation of 20+ models spanning different CIR paradigms, we revealed substantial weaknesses invisible to existing methods: dramatically elevated false-positive rates against explicit negatives, sensitivity to instruction phrasing suggesting overfitting to benchmark patterns, and degraded performance on multi-image reasoning. \textbf{Third} to address these limitations, we explore the usage of a training-free reranking method leveraging off-the-shelf MLLMs that consistently improves performance across all evaluated CIR approaches. \textbf{Finally} PinPoint enables us to benchmark new research directions such as: How can models actively avoid incorrect results? What architectures support multi-image composition? How do we achieve robustness without sacrificing accuracy? Answering these questions will prove key as the field progresses towards human-level visual understanding.





{
    \small
    \bibliographystyle{ieeenat_fullname}
    \bibliography{main}
}


\end{document}